%% file: emnlp2021.tex
\pdfoutput=1

\documentclass[11pt]{article}

\usepackage[]{emnlp2021}

\usepackage{times}
\usepackage{latexsym}
\usepackage{booktabs}
\usepackage{multirow}
\usepackage{threeparttable}
\usepackage{graphicx}

\usepackage[normalem]{ulem}
\useunder{\uline}{\ul}{}
\usepackage[T1]{fontenc}

\usepackage[utf8]{inputenc}

\usepackage{microtype}

\title{VieSum: How Robust Are Transformer-based Models on Vietnamese Summarization?}

\author{Hieu Nguyen$^{1}$, Long Phan$^{1}$, James Anibal$^{2}$, Alec Peltekian$^{1}$, Hieu Tran$^{3,4}$ \\
        $^{1}$Case Western Reserve University \\ $^{2}$National Cancer Institute \\ $^{3}$University of Science, Ho Chi Minh City, Vietnam \\ $^{4}$Vietnam National University, Ho Chi Minh City, Vietnam \\
        \texttt{long.phan@case.edu}}

\begin{document}
\maketitle
\begin{abstract}
\input{section/abstract}
\end{abstract}

\section{Introduction}
\input{section/introduction}

\section{Related works}
\input{section/related_works}

\section{Models}
\input{section/models}

\section{Methods}
\input{section/methods}

\section{Datasets}
\input{section/datasets}
\input{table/results}

\section{Experiments Setup}
\input{section/experments}

\section{Results}
\input{section/results}

\section{Conclusion}
\input{section/conclusion}

\bibliography{anthology,emnlp2021}
\bibliographystyle{acl_natbib}
 
\end{document}

%% file: section/abstract.tex
Text summarization is a challenging task within natural language processing that involves text generation from lengthy input sequences. While this task has been widely studied in English, there is very limited research on summarization for Vietnamese text. In this paper, we investigate the robustness of transformer-based encoder-decoder architectures for Vietnamese abstractive summarization. Leveraging transfer learning and self-supervised learning, we validate the performance of the methods on two Vietnamese datasets.

%% file: section/introduction.tex
In recent years, Transformer-based architecture models and pre-trained language models (LMs) have played an important role in the development of Natural Language Processing (NLP) systems. These large pre-trained models such as ELMo \cite{elmo}, GPT \cite{GPT}, BERT \cite{bert} are trained on a large corpus and have the ability to derive contextual representation of the language(s) in the training data. After pre-training is complete, these models have achieved state-of-the-art results on a broad range of downstream tasks \cite{bert}. These self-supervised learning methods make use of learning objectives such as Masked Language Modeling (MLM) \cite{bert} where random tokens in the input sequence are masked and the model attempts to predict the original tokens in order to gain a better understanding of context. The successes of pre-trained models in English have inspired new research efforts to development pre-trained models in languages such as Vietnamese (i.e., PhoBERT \cite{phobert} and ViBERT\cite{vibert}). There are also ongoing efforts to develop multilingual pre-trained models (mT5 \cite{mT5}, mBART \cite{mbart}) in order to improve performance across multiple languages by learning both general and language-specific representations.

Text summarization is a summarization evaluation task in which an input will be a free-form text paragraph or document(s) and the output sequence will be a short summarization of the input. Although there has been extensive study in plain English text summarization \cite{DBLP:journals/corr/abs-1906-00424} and cross-lingual tasks \cite{DBLP:journals/corr/abs-2012-04307, DBLP:journals/corr/abs-2010-08892}, there have been minimal studies into the Vietnamese text summarization. 

We will perform robust experiments on  Transformer-based encoder-decoder models applied to Vietnamese Summarization Tasks. Following the transformer architecture from \cite{attention} and the Bert2Bert architecture \cite{bert2bert}, we will use existing monolingual Vietnamese pre-trained models and multilingual pre-trained models to perform our experiments on the Vietnamese Summarization task. We will finetune the models on two datasets: Wikilingual \cite{wikilingual} and Vietnews \cite{vietnews}. Our results shows that mT5 and mBART achieve more competitive results than monolingual Vietnamese pretrained model. This indicates that there are significant opportunities to further improve performance on Vietnamese summarization tasks (and Vietnamese NLP in general) with the development of stronger mongolingual pre-trained models.



%% file: section/related_works.tex
There are many abstractive summarization studies in English. In an early example, \cite{gehrmann-etal-2018-bottom} employed a bottom-up content selector (BottomUp) to determine which phrases in the source document should be part of the summary, and then a copy mechanism was applied only to pre-selected phrases during decoding. Their experiments obtained significant improvements on ROUGE for some canonical summarization datasets. 

In recent years, pre-trained language models have been used to enhance performance on language generation tasks. \cite{Liu2019TextSW} developed a Transformer-based encoder-decoder model so pre-trained language models like BERT can be adopted for abstractive summarization task. Here, the authors proposed a novel document-level BERT-based encoder (\textit{BERTSum}) and a general framework encompassing both extractive and abstractive summarization tasks. Based on \textit{BERTSum}, \citet{dou2021gsum} introduced \textit{GSum} that effectively used different types of guidance signals as input in order to generate more suitable words and more accurate summaries. This model accomplished state-of-the-art performance on four popular English summarization datasets.

Meanwhile, there are a small number of studies on Vietnamese text summarization. Most of these focus on inspecting extractive summarization. The researchers \cite{8573420} compared a wide range of extractive methods, including unsupervised ranking methods (e.g., LexRank, LSA, KL-divergence), supervised learning methods using TF-IDF and classifiers (e.g., Support Vector Machine, AdaBoost, Learning-2-rank), and deep learning methods (e.g., Convolutional Neural Network, Long-Short Term Memory). Similarly, the authors \cite{vietnews} also evaluated the extractive methods on their own dataset, which was released publicly as a benchmark for future studies.

Recent work \cite{quoc2021monolingual} investigated the combination of pre-trained BERT model and an unsupervised K-means clustering algorithm on extractive text summarization. The authors utilized multilingual and monolingual BERT models to encode sentence-level contextual information and then ranked this information using K-means algorithm. Their report showed that monolingual models achieved better results compared when to multilingual models performing the same extractive summarization tasks. However, due to the lack of studies on Vietnamese abstract summarization, we compare both multilingual and monolingual encoder-decoder models. 

%% file: section/models.tex
\subsection{BERT}
BERT \cite{bert} (Bidirectional Encoder Representations from Transformers) is a multilayer bidirectional Transformer encoder based on the architecture described in \cite{attention}. In BERT, each input sequence is represented by a sum of the  token embeddings, segment embeddings, its positional embeddings. BERT uses the WordPiece tokenizer to split words into subwords.

\subsection{PhoBERT}
PhoBERT \cite{phobert} is the first public large-scale mongolingual language model pre-trained for Vietnamese. PhoBERT follows the architecture proposed by \cite{roberta}. The PhoBERT pretraining approach is also based on RoBERTa \cite{roberta}, employing the implementation in \textit{fairseq} \cite{fairseq} which optimizes the BERT pre-training procedure for more robust performance. The model is trained on a 20GB word-level Vietnamese news corpus with maximum sequence length of 256.

\subsection{ViBERT}
ViBERT \cite{vibert} leverages checkpoints from the multilingual pre-trained model mBERT \cite{bert} and continues training on monolingual Vietnamese corpus (10GB of syllable-level text). The authors also remove insignificant vocab from mBERT, keeping only Vietnamese vocab.

\subsection{mBERT}
mBERT followings the same training procedure of BERT \cite{bert} on a multilingual dataset. The training languages are the 100 languages with the largest amount of text on Wikipedia, which includes Vietnamese \footnote{https://meta.wikimedia.org/wiki/List\_of\_Wikipedias}.

\subsection{mBART}
mBART \cite{mbart} is a sequence-to-sequence denoising encoder-decoder model that was pre-trained on large-scale
monolingual corpora (including Vietnamese) using the BART objective anđ architecture \cite{bart}.

\subsection{mT5}
mT5 \cite{mT5} is a multilingual variant of the large encoder-decoder pre-trained text-to-text models T5 \cite{t5}. 
mT5 inherits all of the benefits of T5 including the scale and the design which was based on a large-scale empirical study \cite{t5}.

%% file: section/methods.tex
Following the works descried in \cite{attention}, a transformer-based sequence-to-sequence architecture is an encoder-decoder architecture that has multi-layer self-attention. This architecture consists of two components: an encoder and a decoder.

Therefore, in order to generate a target sequence from an input sequence, a full transformer architecture with both an encoder and a decoder is required. PhoBERT and ViBERT are encoder-only models developed for encoding Vietnamese Language representations, so we can quickly assemble these models into the architecture. However, for the decoder level, we change the self-attention layers from bidirectional into left-context-only. We also insert a randomly initialized cross-attention mechanism into the decoder. We denote these models as PhoBERT2PhoBERT and ViBERT2ViBERT. Following the practice from \cite{press-wolf-2017-using}, we tie the input embedding and output embedding in the decoder block, allowing the model to learn from the input embedding weights instead of random weights.


For existing encoder-decoder models like mT5 and mBART, we continuously fine-tune on the released checkpoints while strictly keeping their structure and vocabulary.


%% file: section/datasets.tex
\input{table/data}
We test the VieSum models on two datasets: Wikilingua \cite{wikilingual} and Vietnews \cite{vietnews}. The statistics for these datasets are shown in Table \ref{data}.

\subsection{Wikilingua}
Wikilingua \cite{wikilingual} is a large-scale corpus multilingual corpus for abstractive summarization tasks. The corpus consists of 18 languages, including Vietnamese. These articles and summary pairs are extracted from WikiHow \footnote{https://www.wikihow.com}. 
Wikihow is an online resource containing guidelines and how-to articles for a wide-range of topics. These articles are written by human authors; to ensure quality and accuracy, human experts reviewed the content before inclusion in this study \footnote{https://www.wikihow.com/Experts}. Most of the non-English articles on the site are manually translated from the original English articles. Prior to publication, these foreign language articles are then reviewed by the WikiHow’s international translation team. 

\subsection{Vietnews}
Vietnews \cite{vietnews} collects news data from 3 well-known Vietnamese news websites: \textit{tuoitre.vn}, \textit{vnexpress.net}, and \textit{nguoiduatin.vn}. The collected articles were written from 2016 to 2019. Then the authors remove all articles related to questionnaires, analytical comments and weather forecasts as they are not relevant to document summarization. The final corpus only contains news events.

%% file: table/data.tex
\begin{table*}[]
\centering
\caption{Data statistics of the finetune datasets}
\begin{tabular}{|c|c|l|l|c|l|l}
\hline
\multicolumn{1}{|l|}{}               & \multicolumn{3}{c|}{Wikilingua}          & \multicolumn{3}{c|}{Vietnews}                                    \\ \hline
\multirow{2}{*}{Size}               & \multicolumn{1}{l|}{Train} & Dev  & Test & \multicolumn{1}{l|}{Train}  & Dev   & \multicolumn{1}{l|}{Test}  \\ \cline{2-7} 
                                    & \multicolumn{1}{l|}{13707} & 1957 & 3916 & \multicolumn{1}{l|}{105418} & 22642 & \multicolumn{1}{l|}{22643} \\ \hline
\#avg of words in body     & \multicolumn{3}{c|}{521}                 & \multicolumn{3}{c|}{519}                                         \\ \hline
\#avg of words in abstract & \multicolumn{3}{c|}{44}                  & \multicolumn{3}{c|}{38}                                          \\ \hline
\end{tabular}
\label{data}
\end{table*}

%% file: table/results.tex
\begin{table*}[]
\centering
\caption{Test result on Wikilingua and Vietnews}

\begin{tabular}{l|lll|lll}
\hline
\multirow{2}{*}{Models}                                         & \multicolumn{3}{c|}{WikiLingua}                 & \multicolumn{3}{c|}{Vietnews}                    \\ \cline{2-7} 
                                                                & Rouge-1        & Rouge-2        & Rouge-L       & Rouge-1        & Rouge-2        & Rouge-L        \\ \hline
\begin{tabular}[c]{@{}l@{}}Transformer\\ (RND2RND)\end{tabular} & 46.25          & 16.57          & 29.82         &              57.56  &        24.25        & 35.53          \\ \hline
PhoBERT2RND                                                     & 46.72          & 17.00          & 30.13         & 57.6           & 24.18          & 35.5           \\ \hline
ViBERT2ViBERT                                                   & 53.08          & 20.18    & 31.79         & 59.75          & 27.29 & 36.79          \\ \hline
PhoBERT2PhoBERT                                                 & 50.4     & 19.88          & 32.49   & \textbf{60.37} & \textbf{29.12}          & \textbf{39.44} \\ \hline
mBERT                                                  &       52.82         &        20.57        & 31.55         &       59.67         &    27.36    & 36.73          \\ \hline
mBART                                                           &        {\ul 55.21}        &   {\ul 25.69}             &       {\ul 37.33}        &    {\ul 59.81}            &    {\ul 28.28}            &    {\ul 38.71}            \\ \hline
mT5                                                             & \textbf{55.27} & \textbf{27.63} & \textbf{38.3} & 58.05    & 26.76          & 37.38    \\ \hline
\end{tabular}

\begin{tablenotes}
      \small
      \item \textit{Notes:} The best scores are in bold and second best scores are underlined.
\end{tablenotes}
\label{result}
\end{table*}

%% file: section/experments.tex
\subsection{Baselines}
We verify the effectiveness of our proposed methods by comparing with the Transformer model architecture based on \cite{attention}.
The Transformer models has an encoder and a decoder, in which each layer of encoder-dencoder includes two major components: a multi-head self-attention and a feed-forward network. These model are intialized with random weights and labels RND.

\subsection{Metrics}
ROUGE (Recall-Oriented Understudy for Gisting Evaluation) measures the number of overlapping units (including n-grams and word sequences) between the generated summary and the reference summary.

\begin{itemize}
    \item \textbf{ROUGE-N: } measures overlap between unigrams, bigrams, trigrams and higher order n-grams For our experiments, we will use ROUGE-1 and ROUGE-2 (unigrams and bigrams)
    \item \textbf{ROUGE-L: } measures longest matching sequence of words using Longest Common Subsequences (LCS) with the assumption that longer LCS between the generated summary and the reference summary shows more similarity (higher quality of results).
\end{itemize}

%% file: section/results.tex
We report the results of transformer encoder-decoder models on two datasets: Wikilingua and Vietnews in Table \ref{result}.

\subsection{Wikilingua}
\label{wikilingua}
For the Wikilingua dataset, a first obvious take away is that a pre-trained decoder is important for a transformer model to perform well in Vietnamese Summarization tasks. PhoBERT2RND has minor improvement comparing to untrained Transformer baseline model. Yet, incorporating a pre-trained decoder in PhoBERT2PhoBERT improves the understanding of a transformer model on a Vietnamese language and Summarization. The score improved from 2-4\% accross for all metrics (Rouge-1, Rouge-2, and Rouge-L). Therefore, this setting can be further studied and improved by incorporating a larger pre-trained decoder like GPT \cite{GPT} on Vietnamese language with existing pre-trained encoders.

 mT5 has the highest score across all metrics (Rouge-1, Rouge-2, and Rouge-L), following by mBART model. While PhoBERT2PhoBERT, which was trained monolingual on Vietnamese, has the lowest score compared to other pre-trained encoder-decoder models. This result can be attributed to a training data factor. Both mT5 and mBART train on mC4 \cite{t5} and CC25 \cite{bart} respectively, which are a extracted from a large Common Crawl corpus (\cite{commoncrawl}) \footnote{https://commoncrawl.org/}. Common Crawl is a publicly web archive that provides text scraped from websites. These datasets (C4, CC25, and Common Crawl) may include Vietnamese Wikipedia languages website that help the model excel in constrained domain corpus (Wikipedia). We will further discuss the affect of pre-training data for constrained domain corpus in Section \ref{vietnews}.
 
 \subsection{Vietnews}
 Following the discussion of constrain domain training data in Section \ref{wikilingua}, PhoBERT2PhoBERT, which was trained on 20GB of news text, excels on the Vietnews corpus. On the other hand, large multingual model mT5 and mBART don't have a significant increase in scores compared to RND2RND baseline.
 
 The importance of a pre-trained decoder also shows in the Vietnews corpus. PhoBERT2RND shows approximately the same result as the Transformer baseline.

 \label{vietnews}

%% file: section/conclusion.tex
In this manuscript, we test a series of availbale pre-trained Transformer on Vietnamese Abstractive Summarization Tasks. Following the transformer architecture proposed by \cite{attention}, we showed that incorporating a pre-trained decoder to existed pre-trained encoder will improve the understanding of the model on the Vietnamese language and Summarization ability. We also show that pre-training data quality will affect the performance of the model on a constrain domain.